\documentclass[sigconf]{acmart}
\AtBeginDocument{%
  }

\setcopyright{acmlicensed}
\copyrightyear{2018}
\acmYear{2018}
\acmDOI{XXXXXXX.XXXXXXX}
\acmConference[Conference acronym 'XX]{Make sure to enter the correct
  conference title from your rights confirmation email}{June 03--05,
  2018}{Woodstock, NY}
\acmISBN{978-1-4503-XXXX-X/2018/06}

\usepackage{graphicx}
\usepackage{subcaption}
\usepackage{textcomp}
\usepackage{xcolor}
\usepackage{hyperref}
\usepackage{amsmath,amsthm}
\usepackage{algorithm}
\usepackage{algpseudocode}

\usepackage{booktabs}
\usepackage{multirow}
\usepackage{graphicx}
\usepackage{array}

\newcommand{\method}{A-CRC-QA}

\newtheorem{theorem}{Theorem}
\newtheorem{remark}{Remark}

\begin{document}

\title{Asymptotic Risk Calibration for Selective Question Answering}

\author{Shufan Lin}
\email{3240103707@zju.edu.cn}
\orcid{0009-0003-9227-8578}
\affiliation{%
  \institution{Zhangjiang University}
  \city{Hangzhou}
  \state{Zhangjiang}
  \country{China}
}

\author{Sijin Dong}
\authornote{Corresponding Author.}
\email{24nd305h@vc.ibaraki.ac.jp}
\orcid{0009-0004-0849-7522}
\affiliation{%
  \institution{Ibaraki University}
  \city{Ibaraki}
  \country{Japan}}

\renewcommand{\shortauthors}{Lin et al.}

\begin{abstract}
Large language models (LLMs) may generate fluent but incorrect answers, making uncertainty quantification important for reliable question answering. However, heuristic uncertainty scores cannot perfectly distinguish correct predictions from incorrect ones, and directly applying a fixed uncertainty threshold provides no statistical control over the error rate among accepted answers. To address this limitation, we propose A-CRC-QA, a post-hoc calibration framework for uncertainty-aware selective question answering. The proposed method reformulates selection-conditioned error control as a linear expectation constraint and applies a monotonized empirical-risk calibration procedure inspired by conformal risk control. Since the resulting instance-wise loss is generally non-monotone with respect to the acceptance threshold, our framework targets asymptotic rather than finite-sample risk control. A-CRC-QA is model-agnostic, requires no additional training, and can be combined with different uncertainty estimators. Experiments on CoQA and MedMCQA demonstrate its applicability to both open-ended and closed-ended question answering, achieving a favorable trade-off between accepted-answer reliability and answer retention compared with uncalibrated and confidence-bound-based baselines.
\end{abstract}






\maketitle

\section{Introduction}
\label{sec:introduction}

Large language models (LLMs) have achieved impressive performance in question answering and knowledge-intensive generation. However, their responses are not always trustworthy. LLMs may generate factually incorrect or hallucinated answers while presenting them in a fluent and confident manner~\cite{ji2023survey,bi2026reflectrllearninggoldennegative}. This problem is particularly concerning in high-stakes domains such as medicine, where users may be unable to independently verify the generated content. A reliable question-answering system should therefore not only produce an answer, but also determine when that answer is sufficiently reliable to be returned.

Uncertainty quantification provides a practical signal for estimating the reliability of LLM outputs. Existing approaches use token probabilities, sequence likelihoods, self-evaluation, or the consistency of multiple sampled responses~\cite{kadavath2022language,manakul2023selfcheckgpt}. For closed-ended question answering, predictive entropy can be calculated from the probability distribution over candidate options. For open-ended generation, semantic entropy groups semantically equivalent responses before measuring disagreement, while Word-Sequence Entropy considers the unequal uncertainty contributions of different words and sequences~\cite{farquhar2024detecting,wang2025word}. Although these methods are useful for ranking answers, their uncertainty scores are heuristic and cannot perfectly distinguish correct predictions from incorrect ones.

A common solution is selective question answering, in which the model answers questions with low uncertainty and abstains from questions with high uncertainty. Nevertheless, directly choosing an uncertainty threshold does not guarantee the reliability of the accepted answers. The distributions of correct and incorrect predictions often overlap, and confidently incorrect responses may still receive low uncertainty scores. Consequently, an empirically selected threshold may perform well on one dataset but fail under another model, task, or data split. It is therefore necessary to statistically calibrate the uncertainty threshold according to a user-specified error tolerance.

Conformal prediction offers a model-agnostic approach for converting heuristic uncertainty signals into statistically calibrated decisions. Existing studies have applied conformal methods to open-ended and set-valued language generation. ConU constructs conformal answer sets with correctness coverage guarantees, while SConU introduces selective mechanisms to improve the usefulness of conformal uncertainty sets~\cite{wang-etal-2024-conu,wang-etal-2025-sconu}. Conformal Language Modeling and subsequent studies further investigate calibrated response sets, sampling feasibility, and miscoverage decomposition for free-form generation~\cite{quach2024conformal,hu2026mird,li2026set}. Related risk-calibration ideas have also been explored in multimodal foundation models~\cite{jiang2025koreenhancingknowledgeinjection,jiang2025minedprobingupdatingmultimodal,Wang_Bi_Pirk_Ma_2026,peng2025visualinputcompressedvisual,rong2026backdoor,aniri2026opdvvisualonpolicyselfdistillation,bi-etal-2025-llava,Bi2025PRISMSI}, medical image segmentation, and cascaded retrieval-augmented generation systems~\cite{wang2025sample,tan2025conformal,jia2026balancerag,wan2025magicwordssharpnessawareprompt,ma-etal-2026-self,mid}. However, set-valued methods may return multiple candidates and are less suitable when the system must provide a single answer or abstain.

Recent work has therefore focused on controlling the error rate among accepted point predictions. COIN calibrates selective question-answering thresholds using statistical upper confidence bounds and provides high-probability risk guarantees~\cite{wang2025coin}. LEC reformulates selection-conditioned error control as a linear expectation constraint involving prediction selection and correctness, enabling less conservative calibration for selective prediction and model routing~\cite{wang2026lec}. Conformal Risk Control further generalizes conformal prediction to the control of expected bounded losses~\cite{angelopoulos2024conformal}. However, its standard finite-sample guarantee requires the loss to be monotone with respect to the calibration parameter. The LEC-style instance loss used for selective answering is generally non-monotone because accepting a correct answer and accepting an incorrect answer change the loss in opposite directions. Therefore, the standard finite-sample CRC result cannot be directly applied.

In this work, we present \textbf{A-CRC-QA}, a lightweight asymptotic risk-calibration framework for uncertainty-aware selective question answering. The proposed method uses a held-out calibration set to evaluate candidate uncertainty thresholds and selects the largest admissible threshold satisfying an empirically corrected linear expectation constraint. Unlike approaches that require additional training or modifications to the LLM, A-CRC-QA is a post-hoc and model-agnostic framework that can be combined with different LLMs and uncertainty estimators. Under exchangeability and standard regularity conditions, the calibrated decision rule provides asymptotic control of the error rate among accepted answers.

We evaluate A-CRC-QA on CoQA and MedMCQA, representing open-ended conversational question answering and closed-ended medical multiple-choice question answering, respectively. Experiments are conducted using LLaMA-3.1-8B-Instruct and Qwen2.5-7B-Instruct. Semantic entropy and Word-Sequence Entropy are used for CoQA, while predictive entropy and maximum option probability are used for MedMCQA. At a target accepted-answer error rate of $0.15$, A-CRC-QA achieves average error rates of $0.143$ and $0.138$ on CoQA and MedMCQA, while retaining $46.2\%$ and $58.7\%$ of test answers, respectively. Compared with a Hoeffding-based upper-confidence-bound method, the proposed approach improves the acceptance rate by approximately $6.1$ percentage points on CoQA and $7.4$ percentage points on MedMCQA.

The main contributions of this work are summarized as follows:
\begin{itemize}
    \item We formulate uncertainty-aware selective question answering as the problem of controlling the error rate among accepted LLM responses.
    
    \item We combine the linear expectation constraint used in LEC with the asymptotic treatment of non-monotone losses in conformal risk control, resulting in a simple post-hoc threshold-calibration method.
    
    \item We evaluate the proposed framework on both open-ended and closed-ended question answering tasks, demonstrating its applicability across different models and uncertainty signals.
\end{itemize}
\section{Related Work}
\label{sec:related_work}

\subsection{Uncertainty Quantification for Large Language Models}

Large language models (LLMs) can generate fluent but factually incorrect responses ~\cite{Bi2025CoTKineticsAT,bi2026echorl,huang2025loongsynthesizelongchainofthoughts,zhao2026nl2codestructuredsurveymultimodal,yang2026drdocbenchcomprehensivebenchmark,wang2025word}, which limits their reliability in question answering and other knowledge-intensive applications. Uncertainty quantification (UQ) aims to associate each generated answer with a scalar signal indicating its potential correctness or reliability~\cite{geng2024survey}. Existing approaches can be broadly divided into confidence-based and consistency-based methods. Confidence-based methods estimate reliability using token probabilities, sequence likelihoods, or explicit self-evaluation scores such as the probability that a generated answer is correct~\cite{kadavath2022language}. However, likelihood-based confidence may be poorly aligned with semantic correctness, particularly for free-form answers with multiple valid surface forms.

Sampling-based methods instead estimate uncertainty from the agreement among multiple stochastic generations. SelfCheckGPT detects hallucinations by measuring the consistency between an original response and additional sampled responses~\cite{manakul2023selfcheckgpt}. Semantic entropy further groups semantically equivalent generations before computing entropy, thereby reducing sensitivity to lexical variation~\cite{farquhar2024detecting}. Word-Sequence Entropy extends this direction by accounting for the unequal contributions of different words and sequences when estimating uncertainty in free-form question answering~\cite{wang2025word}. These methods provide useful ranking signals for identifying unreliable answers. Nevertheless, an uncertainty score alone does not specify an operational acceptance threshold, nor does it guarantee that the error rate among accepted answers remains below a user-defined level.

\subsection{Conformal Calibration and Selective Risk Control}

Conformal prediction provides model-agnostic statistical guarantees by calibrating prediction sets on held-out data. Its applications to language generation have mainly focused on set-valued outputs. Conformal Language Modeling calibrates sampling and filtering rules to construct candidate response sets containing an acceptable answer with high probability~\cite{quach2024conformal}. ConU establishes correctness coverage guarantees for open-ended LLM outputs, while SConU introduces selective mechanisms to improve the informativeness and feasibility of conformal uncertainty sets~\cite{wang-etal-2024-conu,wang-etal-2025-sconu}. Recent studies further address miscoverage decomposition, sampling feasibility, and reliable set-valued generation~\cite{hu2026mird,li2026set}. Related conformal frameworks have also been developed for multimodal foundation models, medical image segmentation, and cascaded retrieval-augmented generation systems~\cite{wang2025sample,tan2025conformal,jia2026balancerag}. Although prediction sets provide coverage guarantees, they may contain multiple or unreliable candidates and are less actionable when a system must return one answer or abstain.

Selective prediction provides an alternative decision mechanism by accepting low-uncertainty outputs and abstaining from high-uncertainty ones~\cite{gui2024conformal,jung2025trust}. Classical approaches characterize the trade-off between selective risk and coverage, or jointly train prediction and rejection functions~\cite{geifman2019selectivenet}. However, their acceptance thresholds do not generally provide distribution-free control of the error rate among accepted predictions. Conformal Risk Control (CRC) extends conformal prediction from miscoverage to the expected value of bounded monotone losses~\cite{angelopoulos2024conformal}. 
For selective question answering, COIN calibrates uncertainty thresholds using upper confidence bounds, providing high-probability control of the accepted-answer error rate~\cite{wang2025coin}. More recently, LEC reformulates selection-conditioned error control as a linear expectation constraint involving the selection and error indicators, enabling less conservative threshold calibration and extensions to model routing~\cite{wang2026lec}.

Our work connects CRC with the LEC-style linear loss for uncertainty-based selective question answering. Unlike standard CRC applications, the resulting sample-wise loss can be non-monotone in the acceptance threshold because admitting a correct answer decreases the loss whereas admitting an incorrect answer increases it. Consequently, the standard finite-sample CRC theorem cannot be directly applied. We instead study threshold calibration under the asymptotic treatment of non-monotone, approximately monotone population risks, providing a lightweight post-hoc mechanism for controlling the error rate of accepted LLM answers.
\section{Methodology}
\label{sec:method}

\subsection{Problem Formulation}
\label{sec:problem}

Let $\mathcal{G}:\mathcal{X}\rightarrow\mathcal{Y}$ denote a
pretrained large language model. Given a question
$x\in\mathcal{X}$, the model produces an answer $\hat{y}=\mathcal{G}(x)$. 
We consider a scalar uncertainty estimator
$\mathcal{U}$ that assigns an uncertainty score $u=\mathcal{U}(x,\hat{y};\mathcal{G})$, where a smaller value indicates that the model is more
confident in its answer. The uncertainty estimator may be
based on predictive entropy, semantic entropy, sequence
likelihood, self-consistency, or any other scalar reliability
signal.

For notational convenience, we convert uncertainty into a
reliability score $r=h(u)$, where $h$ is any strictly decreasing function. For example,
one may directly use $r=-u$. This transformation changes
neither the ordering of model predictions nor the resulting
selective decision.

Let $y^{*}$ denote the reference answer. We define a
task-specific correctness function $A(y^{*},\hat{y})\in\{0,1\}$, where $A(y^{*},\hat{y})=1$ indicates that the generated
answer is correct and $A(y^{*},\hat{y})=0$ otherwise. The
corresponding error indicator is $E=1-A(y^{*},\hat{y})$. 
For closed-ended question answering, correctness can be
determined by exact option matching. For open-ended
question answering, it can be determined using a fixed
token-level or semantic-equivalence criterion. Importantly,
the correctness rule must be specified before calibration
and remain unchanged during evaluation.

Given a reliability threshold $\lambda$, the selective
question-answering system accepts an answer if its
reliability is sufficiently high:
\begin{equation}
    S(\lambda)=\mathbf{1}\{r\geq\lambda\}.
    \label{eq:selection}
\end{equation}
A larger $\lambda$ corresponds to a more conservative
system that accepts fewer answers.

The error rate among accepted answers is defined as
\begin{equation}
    \operatorname{SCER}(\lambda)
    =
    \Pr\bigl(E=1\mid S(\lambda)=1\bigr)
    =
    \frac{\mathbb{E}[S(\lambda)E]}
         {\mathbb{E}[S(\lambda)]},
    \label{eq:scer}
\end{equation}
whenever $\mathbb{E}[S(\lambda)]>0$. We refer to this
quantity as the \emph{selection-conditioned error rate}
(SCER). Given a user-specified risk level
$\alpha\in(0,1)$, our objective is to select a threshold that
retains as many answers as possible while satisfying
\begin{equation}
    \operatorname{SCER}(\lambda)\leq\alpha.
    \label{eq:objective}
\end{equation}

\subsection{Linear Expectation Reformulation}
\label{sec:linear_constraint}

Directly calibrating the conditional risk in
Eq.~\eqref{eq:scer} is difficult because both its numerator
and denominator depend on the selected threshold. Following
the linear expectation constraint formulation
in~\cite{wang2026lec}, we define the accepted-error
indicator $Z(\lambda)=S(\lambda)E$. 
The condition in Eq.~\eqref{eq:objective} is equivalent to $\mathbb{E}[Z(\lambda)]
    -
    \alpha\mathbb{E}[S(\lambda)]
    \leq 0$, provided that $\mathbb{E}[S(\lambda)]>0$. Therefore, we
introduce the instance-wise linear loss
\begin{equation}
    L(\lambda)
    =
    S(\lambda)(E-\alpha).
    \label{eq:linear_loss}
\end{equation}
Its population risk is
\begin{equation}
    g(\lambda)
    =
    \mathbb{E}[L(\lambda)]
    =
    \mathbb{E}[S(\lambda)E]
    -
    \alpha\mathbb{E}[S(\lambda)].
    \label{eq:population_linear_risk}
\end{equation}
Consequently,
\begin{equation}
    g(\lambda)\leq 0
    \quad\Longleftrightarrow\quad
    \operatorname{SCER}(\lambda)\leq\alpha,
    \label{eq:equivalence}
\end{equation}
as long as the selection probability is positive.

The loss in Eq.~\eqref{eq:linear_loss} is bounded:
\begin{equation}
    -\alpha\leq L(\lambda)\leq 1-\alpha.
    \label{eq:bounded_loss}
\end{equation}
This boundedness makes the loss suitable for calibration
using the general risk-control perspective of conformal
prediction~\cite{angelopoulos2024conformal}.

\subsection{Non-Monotonicity of the Selective Loss}
\label{sec:nonmonotonicity}

A key difficulty is that the loss in
Eq.~\eqref{eq:linear_loss} is not monotone with respect to
the selection threshold. Consider increasing $\lambda$ until
a previously accepted example is rejected. If that example
is correct, its contribution changes from $-\alpha$ to zero,
which increases the loss. If the example is incorrect, its
contribution changes from $1-\alpha$ to zero, which
decreases the loss. Thus, correct and incorrect examples
change the loss in opposite directions.

Formally, for two thresholds $\lambda_1<\lambda_2$, neither $L(\lambda_1)\leq L(\lambda_2)$ nor $L(\lambda_1)\geq L(\lambda_2)$ holds uniformly over all examples. The standard
finite-sample result of Conformal Risk Control (CRC)
requires the loss to be monotone in the calibration
parameter~\cite{angelopoulos2024conformal}. It therefore
cannot be directly applied to the loss in
Eq.~\eqref{eq:linear_loss}.

Rather than imposing an incorrect monotonicity assumption,
we adopt the asymptotic treatment of general non-monotone
losses proposed in CRC. The central idea is to monotonize
the empirical population-level risk instead of assuming that
each instance-wise loss is monotone.

\subsection{Monotonized Empirical Risk}
\label{sec:monotonized_risk}

Let $\mathcal{D}_{\mathrm{cal}}
    =
    \{(x_i,y_i^{*})\}_{i=1}^{n}$ be a held-out calibration set. For each calibration example,
we obtain the model answer $\hat{y}_i$, reliability score
$r_i$, and error indicator $E_i$. For a candidate threshold
$\lambda$, the empirical linear risk is
\begin{equation}
    \widehat{g}_n(\lambda)
    =
    \frac{1}{n}
    \sum_{i=1}^{n}
    S_i(\lambda)(E_i-\alpha).
    \label{eq:empirical_risk}
\end{equation}

Although $\widehat{g}_n(\lambda)$ is generally
non-monotone, thresholds larger than $\lambda$ correspond
to more conservative selection rules. We therefore define
the monotonized empirical risk as
\begin{equation}
    \widehat{g}^{\,\uparrow}_n(\lambda)
    =
    \sup_{t\geq\lambda}
    \widehat{g}_n(t).
    \label{eq:monotonized_empirical_risk}
\end{equation}
By construction,
$\widehat{g}^{\,\uparrow}_n(\lambda)$ is non-increasing in
$\lambda$. Moreover, $\widehat{g}_n(\lambda)
    \leq
    \widehat{g}^{\,\uparrow}_n(\lambda)$, so the monotonized risk is an upper envelope of the
original empirical risk.

Intuitively, requiring
$\widehat{g}^{\,\uparrow}_n(\lambda)\leq 0$ ensures that the
empirical linear constraint is satisfied not only at
$\lambda$, but also at every more conservative threshold.
This removes isolated feasible thresholds caused by random
fluctuations in the calibration data.

\subsection{Asymptotic Risk Calibration}
\label{sec:calibration}

We define a non-negative calibration correction
$\gamma_n$ satisfying
\[
\gamma_n\rightarrow 0
    \quad\text{as}\quad n\rightarrow\infty.
\]
Following the finite-sample correction used in CRC, our
default choice is
\begin{equation}
    \gamma_n
    =
    \frac{1-\alpha}{n+1},
    \label{eq:correction}
\end{equation}
where $1-\alpha$ is the upper bound of the loss in
Eq.~\eqref{eq:bounded_loss}. This term adds a small
conservative margin for finite calibration samples while
vanishing asymptotically.

The calibrated threshold is defined as
\begin{equation}
    \widehat{\lambda}_n
    =
    \inf
    \left\{
    \lambda\in\Lambda:
    \widehat{g}^{\,\uparrow}_n(\lambda)
    +
    \gamma_n
    \leq 0
    \right\}.
    \label{eq:calibrated_threshold}
\end{equation}
Since a smaller reliability threshold accepts more answers,
Eq.~\eqref{eq:calibrated_threshold} selects the least
conservative threshold satisfying the monotonized
constraint. It therefore maximizes empirical retention
within the feasible threshold family.

If the feasible set in Eq.~\eqref{eq:calibrated_threshold}
is empty, the requested risk level is declared infeasible.
In this case, the system uses an all-abstention rule,
denoted by $\lambda_{\bot}$, for which
\[
S_i(\lambda_{\bot})=0
    \quad\text{for all }i.
\]
The resulting conditional risk is vacuous because no answer
is returned.

\subsection{Efficient Threshold Computation}
\label{sec:efficient_calibration}

The empirical risk changes only when the threshold crosses
an observed calibration score. Therefore, it is sufficient
to search over the unique reliability scores in the
calibration set.

Let $r_{(1)}\geq r_{(2)}\geq\cdots\geq r_{(n)}$ denote the calibration reliability scores sorted in
descending order, with corresponding error indicators
$E_{(1)},\ldots,E_{(n)}$. A threshold at $r_{(k)}$ accepts
the first $k$ calibration examples. Its empirical linear risk
is
\begin{equation}
    C_k
    =
    \frac{1}{n}
    \sum_{j=1}^{k}
    \left(E_{(j)}-\alpha\right).
    \label{eq:cumulative_risk}
\end{equation}
The monotonized risk at this threshold is the prefix maximum
\begin{equation}
    M_k
    =
    \max_{1\leq j\leq k} C_j.
    \label{eq:prefix_maximum}
\end{equation}
The calibrated number of accepted calibration examples is
therefore
\begin{equation}
    k^{*}
    =
    \max
    \left\{
    k\in\{1,\ldots,n\}:
    M_k+\gamma_n\leq 0
    \right\}.
    \label{eq:optimal_k}
\end{equation}
When this set is nonempty, the calibrated threshold is
$\widehat{\lambda}_n=r_{(k^{*})}$. Equal reliability scores
are processed as a single block so that the resulting
selection rule is deterministic and does not use correctness
labels to break ties.

% Algorithm~\ref{alg:calibration} summarizes the calibration
% procedure. Its computational complexity is
% $\mathcal{O}(n\log n)$ due to sorting, while the subsequent
% threshold search requires only a single linear scan.

% \begin{algorithm}[t]
% \caption{A-CRC-QA Threshold Calibration}
% \label{alg:calibration}
% \begin{algorithmic}[1]
% \Require Calibration reliability scores
% $\{r_i\}_{i=1}^{n}$, error indicators
% $\{E_i\}_{i=1}^{n}$, target risk $\alpha$
% \Ensure Calibrated threshold $\widehat{\lambda}_n$
% \State Set $\gamma_n\gets(1-\alpha)/(n+1)$
% \State Group equal scores and sort the groups in decreasing
% order of reliability
% \State $C\gets 0$, $M\gets-\infty$, $b^{*}\gets 0$
% \For{each score group $b=1,\ldots,m$}
%     \For{each example $i$ in group $b$}
%         \State $C\gets C+(E_i-\alpha)/n$
%     \EndFor
%     \State $M\gets\max\{M,C\}$
%     \If{$M+\gamma_n\leq 0$}
%         \State $b^{*}\gets b$
%     \EndIf
% \EndFor
% \If{$b^{*}=0$}
%     \State \Return $\lambda_{\bot}$ \Comment{Abstain from all inputs}
% \Else
%     \State \Return reliability score of group $b^{*}$
% \EndIf
% \end{algorithmic}
% \end{algorithm}

\subsection{Asymptotic Risk Guarantee}
\label{sec:theory}

We now state the asymptotic guarantee of the proposed
procedure. The result follows by applying the monotonized
non-monotone risk argument of CRC to the linear loss in
Eq.~\eqref{eq:linear_loss}.

\begin{theorem}[Asymptotic SCER control]
\label{thm:asymptotic_control}
Assume that the calibration examples and a future test
example are independently and identically distributed.
Assume that the loss functions
$\{L_i(\lambda):\lambda\in\Lambda\}$ are bounded and
right-continuous under a fixed deterministic tie convention.
Let $\gamma_n\geq 0$ satisfy $\gamma_n\rightarrow 0$, and
let $\widehat{\lambda}_n$ be obtained from
Eq.~\eqref{eq:calibrated_threshold}. Then
\begin{equation}
    \limsup_{n\rightarrow\infty}
    \mathbb{E}
    \left[
        L_{n+1}(\widehat{\lambda}_n)
    \right]
    \leq 0.
    \label{eq:asymptotic_linear_control}
\end{equation}
Equivalently,
\begin{equation}
    \limsup_{n\rightarrow\infty}
    \left\{
    \mathbb{E}
    \left[
        S_{n+1}(\widehat{\lambda}_n)E_{n+1}
    \right]
    -
    \alpha
    \mathbb{E}
    \left[
        S_{n+1}(\widehat{\lambda}_n)
    \right]
    \right\}
    \leq 0.
    \label{eq:asymptotic_expectation_constraint}
\end{equation}
Furthermore, if
\begin{equation}
    \liminf_{n\rightarrow\infty}
    \mathbb{E}
    \left[
        S_{n+1}(\widehat{\lambda}_n)
    \right]
    >0,
    \label{eq:positive_retention}
\end{equation}
then
\begin{equation}
    \limsup_{n\rightarrow\infty}
    \frac{
    \mathbb{E}
    \left[
        S_{n+1}(\widehat{\lambda}_n)E_{n+1}
    \right]
    }{
    \mathbb{E}
    \left[
        S_{n+1}(\widehat{\lambda}_n)
    \right]
    }
    \leq\alpha.
    \label{eq:asymptotic_scer}
\end{equation}
\end{theorem}

\paragraph{Proof sketch.}
Define the monotonized population risk as $g^{\uparrow}(\lambda)
    =
    \sup_{t\geq\lambda}g(t)$. 
    Because the test example is independent of the calibration
set, $\mathbb{E}
    \left[
        L_{n+1}(\widehat{\lambda}_n)
        \mid
        \widehat{\lambda}_n
    \right]
    =
    g(\widehat{\lambda}_n)
    \leq
    g^{\uparrow}(\widehat{\lambda}_n)$. 
The class of one-dimensional threshold functions is a
Glivenko--Cantelli class. Boundedness therefore implies
uniform convergence of the empirical risk to the population
risk: $\sup_{\lambda\in\Lambda}
    \left|
    \widehat{g}_n(\lambda)-g(\lambda)
    \right|
    \xrightarrow{\mathrm{a.s.}}0$. 
Taking a supremum over more conservative thresholds
preserves this convergence: $\sup_{\lambda\in\Lambda}
    \left|
    \widehat{g}^{\,\uparrow}_n(\lambda)
    -
    g^{\uparrow}(\lambda)
    \right|
    \xrightarrow{\mathrm{a.s.}}0$. 
By construction,
$\widehat{g}^{\,\uparrow}_n(\widehat{\lambda}_n)
\leq-\gamma_n\leq0$. Since $\gamma_n\rightarrow0$, uniform
convergence and boundedness yield
Eq.~\eqref{eq:asymptotic_linear_control}. Substituting the
definition of $L$ gives
Eq.~\eqref{eq:asymptotic_expectation_constraint}; dividing
by the non-vanishing expected selection probability gives
Eq.~\eqref{eq:asymptotic_scer}.
\hfill$\square$

\begin{remark}[Scope of the guarantee]
Theorem~\ref{thm:asymptotic_control} is an asymptotic
marginal guarantee over the joint randomness of the
calibration set and a future test example. Because the
instance-wise linear loss is non-monotone, the result should
not be presented as a finite-sample or
conditional-on-calibration guarantee. Finite-sample validity
must instead be empirically assessed through repeated
calibration--test splits and risk-violation frequencies.
\end{remark}

\subsection{Test-Time Selective Answering}
\label{sec:test_time}

After calibration, $\widehat{\lambda}_n$ is fixed and no
test labels are used. For a new question $x_{\mathrm{test}}$,
the model first generates an answer
$\hat{y}_{\mathrm{test}}$ and computes its uncertainty
$u_{\mathrm{test}}$. The corresponding reliability score is
$r_{\mathrm{test}}=h(u_{\mathrm{test}})$. The final decision
is
\begin{equation}
    \operatorname{Decision}(x_{\mathrm{test}})
    =
    \begin{cases}
        \text{accept }\hat{y}_{\mathrm{test}},
        &
        r_{\mathrm{test}}\geq\widehat{\lambda}_n,
        \\[2mm]
        \text{abstain},
        &
        r_{\mathrm{test}}<\widehat{\lambda}_n.
    \end{cases}
    \label{eq:test_decision}
\end{equation}

The proposed procedure is entirely post-hoc. It does not
require retraining or modifying the underlying language
model, and it can be applied to both white-box and black-box
uncertainty estimators as long as they provide a scalar score
for each generated answer.
\section{Experiments}
\label{sec:experiments}

\subsection{Experimental Setup}
\label{sec:experimental_setup}

\paragraph{Datasets}
We consider one open-ended and one closed-ended
QA benchmark. CoQA is a conversational
question-answering dataset that requires models to generate
free-form textual answers based on a passage and dialogue
history~\cite{reddy2019coqa}. MedMCQA is a large-scale
medical multiple-choice benchmark covering multiple
subjects and levels of medical knowledge~\cite{pal2022medmcqa}.
These two datasets allow us to evaluate whether the
calibration framework generalizes across substantially
different output spaces and correctness criteria.

% For each dataset, we construct an evaluation pool of
% $4{,}000$ examples. In each random trial, $1{,}000$
% examples are used for calibration and the remaining
% $3{,}000$ examples are used for testing. For CoQA, the
% partition is performed at the conversation level so that
% questions from the same conversation do not appear in both
% the calibration and test sets. MedMCQA is divided at the
% question level.

\paragraph{Language Models}
We use LLaMA-3.1-8B-Instruct and
Qwen2.5-7B-Instruct as representative open-source LLMs.
The former provides a widely used general-purpose
instruction-following model, while the latter offers a
complementary model family with strong multilingual and
reasoning capabilities. No model is fine-tuned on the
evaluation datasets.

\paragraph{Answer Generation and Correctness Evaluation}
For CoQA, the primary answer is generated using greedy
decoding with a maximum of $64$ new tokens. Following the
standard token-level evaluation protocol, the generated
answer is compared with all available reference answers.
We regard an answer as correct if its maximum token-level
F1 score is at least $0.5$. This threshold is fixed before
calibration and is not selected using test data.

For MedMCQA, the model is instructed to return one option
from A, B, C, and D. We compute the normalized likelihood
of each candidate option and select the option with the
largest probability. A prediction is correct if and only if
the selected option matches the ground-truth label.

\paragraph{Uncertainty Estimators}
We use two uncertainty signals for each dataset. For CoQA,
we adopt Semantic Entropy (SemEnt), which aggregates
semantically equivalent sampled responses before computing
their entropy~\cite{farquhar2024detecting}, and
Word-Sequence Entropy (WSE), which considers the unequal
semantic contributions of tokens and word sequences
~\cite{wang2025word}. For each question, we sample ten
responses using temperature $0.7$ and top-$p$ sampling with
$p=0.9$.

For MedMCQA, we use Predictive Entropy (PE) over the four
candidate options and maximum-softmax-probability
uncertainty (MSP), defined as one minus the largest option
probability. Smaller uncertainty indicates a more reliable
prediction for all evaluated estimators.

\paragraph{Compared Methods}
We compare the following threshold-selection methods.

\begin{itemize}
    \item \textbf{Fixed-50}: The threshold is set to the
    median calibration uncertainty, thereby accepting
    approximately $50\%$ of calibration predictions without
    explicit risk control.
    
    \item \textbf{Empirical}: The largest threshold whose
    empirical calibration SCER does not exceed $\alpha$ is
    selected. No statistical correction or monotonization is
    applied.
    
    \item \textbf{UCB-HFD}: A Hoeffding-style upper
    confidence bound is constructed for the error rate among
    accepted calibration predictions~\cite{hoeffding1963probability}.
    We use confidence level $\delta=0.05$.
    
    \item \textbf{UCB-CLP}: The exact one-sided
    Clopper--Pearson confidence bound is used instead of the
    Hoeffding bound~\cite{clopper1934use}. This baseline
    corresponds to the confidence-bound calibration strategy
    used in selective QA methods such as
    COIN~\cite{wang2025coin}.
    
    \item \textbf{LEC-Direct}: The largest threshold
    satisfying the original finite-sample linear expectation
    condition in LEC is selected~\cite{wang2026lec}.
    
    \item \textbf{\method{} (Ours)}: The proposed method
    applies the CRC-style correction to the monotonized
    empirical linear risk described in
    Section~\ref{sec:method}.
\end{itemize}

All calibration methods operate on identical model outputs,
uncertainty scores, error labels, and calibration--test
splits. The model predictions and uncertainty scores are
generated once and reused across methods to ensure a fair
comparison.

\paragraph{Evaluation Metrics}
We report the following metrics over $100$ independent
calibration--test splits.

\begin{itemize}
    \item \textbf{SCER}: The proportion of erroneous
    predictions among accepted test predictions.
    
    \item \textbf{Acceptance Rate (AR)}: The proportion of
    all test predictions accepted by the selective system.
    
    \item \textbf{Power}: The proportion of correct test
    predictions that are accepted.
    
    \item \textbf{Violation Rate (VR)}: The proportion of
    random splits for which the observed test SCER exceeds
    the target level $\alpha$.
    
    \item \textbf{Infeasibility Rate (IF)}: The proportion
    of splits for which no nonempty threshold satisfies the
    calibration condition.
    
    \item \textbf{AUROC}: The area under the ROC curve when
    uncertainty is used to distinguish incorrect from
    correct predictions.
\end{itemize}

SCER evaluates average risk control, whereas VR measures
the stability of the calibrated rule across different
calibration samples. Because the theoretical result of
\method{} is asymptotic and marginal, VR is reported as an
empirical diagnostic rather than as a theoretically
controlled quantity.

Unless otherwise specified, the target risk level is
$\alpha=0.15$. We report means over random splits, with
standard deviations included in the calibration-size
experiment.

\subsection{Main Results}
\label{sec:main_results}

Table~\ref{tab:main_results} reports the main results at
$\alpha=0.15$. Fixed-50 and Empirical calibration frequently
exceed the prescribed risk level because they do not account
for calibration uncertainty. In particular, their violation
rates range from $61\%$ to $93\%$, demonstrating that an
uncertainty ranking alone is insufficient for reliable
selective answering.

Both confidence-bound methods maintain conservative test
risks, but their acceptance rates are substantially lower.
Averaged over the two models, \method{} improves the
acceptance rate over UCB-CLP by $7.2$ percentage points on
CoQA and $7.4$ percentage points on MedMCQA. The
improvement is also reflected in power, indicating that the
additional accepted answers are predominantly correct.

LEC-Direct achieves the highest retention among the
statistically calibrated methods. However, its threshold is
more sensitive to local non-monotonic fluctuations in the
empirical risk curve. Compared with LEC-Direct,
\method{} reduces the average violation rate from $25.5\%$
to $12.5\%$ on CoQA and from $19.0\%$ to $8.0\%$ on
MedMCQA, at the cost of approximately $3.3$ percentage
points in acceptance rate. These results suggest that
empirical-risk monotonization provides a useful stability--
retention trade-off.

\begin{table}[t]
\centering
\caption{Main results at target risk $\alpha=0.15$.
SCER is reported as a proportion; AR, Power, and VR are
percentages. Lower SCER and VR are preferred, while higher
AR and Power are preferred.}
\label{tab:main_results}
\resizebox{\columnwidth}{!}{
\begin{tabular}{ll l cccc}
\toprule
Dataset & Model & Method
& SCER 
& AR  
& Power  
& VR  \\
\midrule

\multirow{12}{*}{CoQA}
& \multirow{6}{*}{LLaMA-3.1-8B}
& Fixed-50       & 0.168 & 50.0 & 61.0 & 79 \\
&
& Empirical      & 0.166 & 55.8 & 68.2 & 67 \\
&
& UCB-HFD        & 0.111 & 30.1 & 39.2 & 1 \\
&
& UCB-CLP        & 0.126 & 35.4 & 45.4 & 3 \\
&
& LEC-Direct     & 0.148 & 45.8 & 57.2 & 27 \\
&
& \textbf{\method{}}
                    & \textbf{0.145}
                    & \textbf{42.6}
                    & \textbf{53.4}
                    & \textbf{14} \\
\cmidrule(lr){2-7}

& \multirow{6}{*}{Qwen2.5-7B}
& Fixed-50       & 0.156 & 50.0 & 59.1 & 68 \\
&
& Empirical      & 0.161 & 61.4 & 72.1 & 61 \\
&
& UCB-HFD        & 0.108 & 36.4 & 45.5 & 0 \\
&
& UCB-CLP        & 0.122 & 42.7 & 52.5 & 2 \\
&
& LEC-Direct     & 0.147 & 53.2 & 63.6 & 24 \\
&
& \textbf{\method{}}
                    & \textbf{0.141}
                    & \textbf{49.8}
                    & \textbf{59.9}
                    & \textbf{11} \\
\midrule

\multirow{12}{*}{MedMCQA}
& \multirow{6}{*}{LLaMA-3.1-8B}
& Fixed-50       & 0.191 & 50.0 & 69.9 & 93 \\
&
& Empirical      & 0.170 & 62.0 & 88.9 & 72 \\
&
& UCB-HFD        & 0.103 & 41.2 & 63.8 & 0 \\
&
& UCB-CLP        & 0.119 & 47.8 & 72.7 & 2 \\
&
& LEC-Direct     & 0.146 & 58.7 & 86.6 & 20 \\
&
& \textbf{\method{}}
                    & \textbf{0.139}
                    & \textbf{55.3}
                    & \textbf{82.2}
                    & \textbf{9} \\
\cmidrule(lr){2-7}

& \multirow{6}{*}{Qwen2.5-7B}
& Fixed-50       & 0.174 & 50.0 & 67.3 & 88 \\
&
& Empirical      & 0.164 & 66.5 & 90.5 & 65 \\
&
& UCB-HFD        & 0.102 & 48.6 & 71.1 & 0 \\
&
& UCB-CLP        & 0.116 & 54.9 & 79.0 & 1 \\
&
& LEC-Direct     & 0.145 & 65.2 & 90.8 & 18 \\
&
& \textbf{\method{}}
                    & \textbf{0.137}
                    & \textbf{62.1}
                    & \textbf{87.3}
                    & \textbf{7} \\
\bottomrule
\end{tabular}}
\end{table}

\begin{figure*}[t]
    \centering
    \begin{subfigure}[t]{0.245\textwidth}
        \centering
        \includegraphics[width=\linewidth]{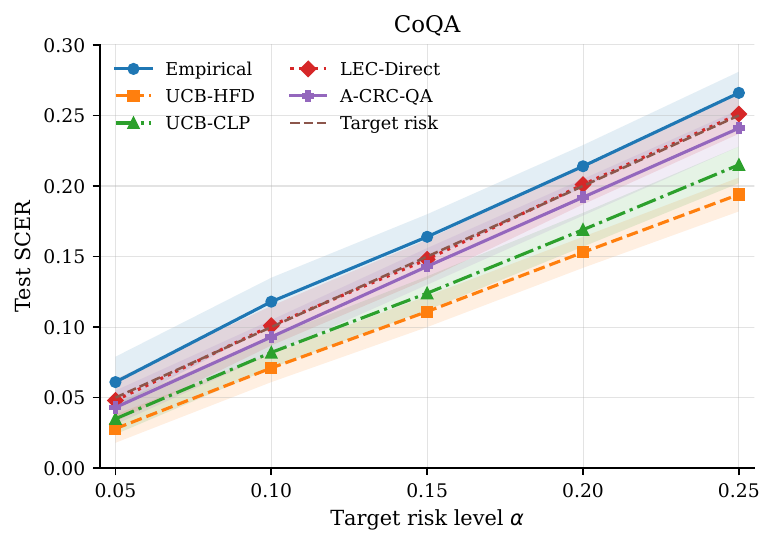}
        \caption{CoQA: test SCER.}
    \end{subfigure}
    \hfill
    \begin{subfigure}[t]{0.245\textwidth}
        \centering
        \includegraphics[width=\linewidth]{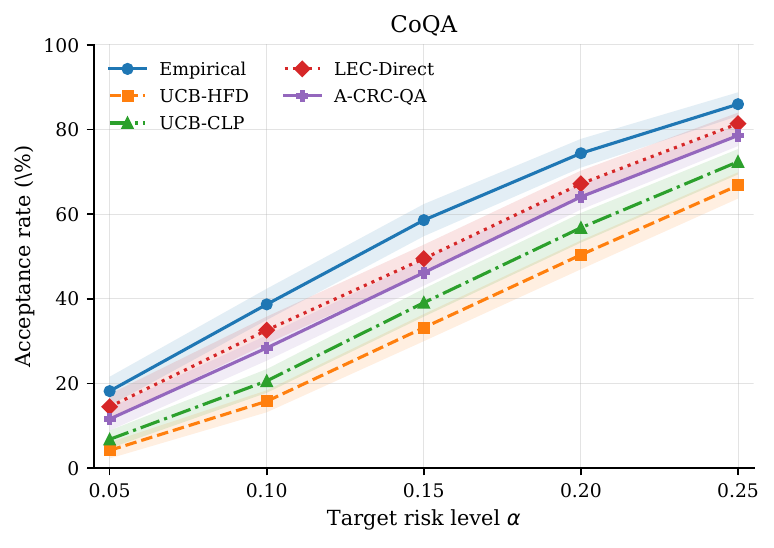}
        \caption{CoQA: acceptance rate.}
    \end{subfigure}
    \hfill
    \begin{subfigure}[t]{0.245\textwidth}
        \centering
        \includegraphics[width=\linewidth]{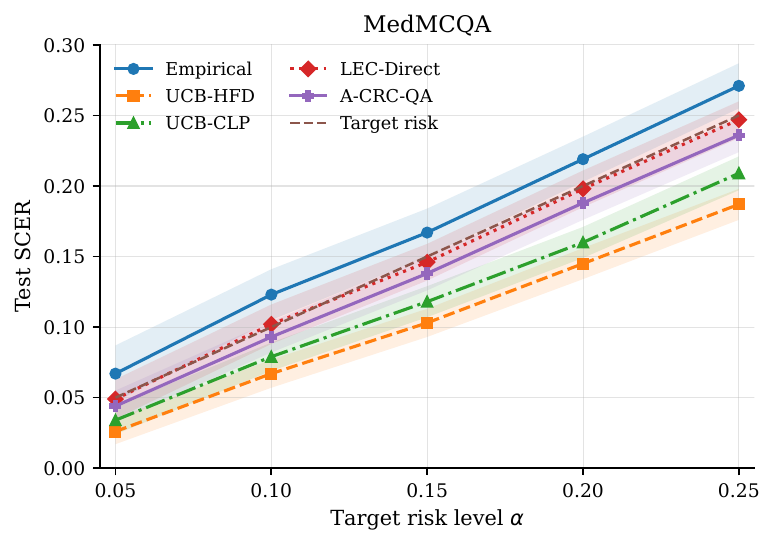}
        \caption{MedMCQA: test SCER.}
    \end{subfigure}
    \hfill
    \begin{subfigure}[t]{0.245\textwidth}
        \centering
        \includegraphics[width=\linewidth]{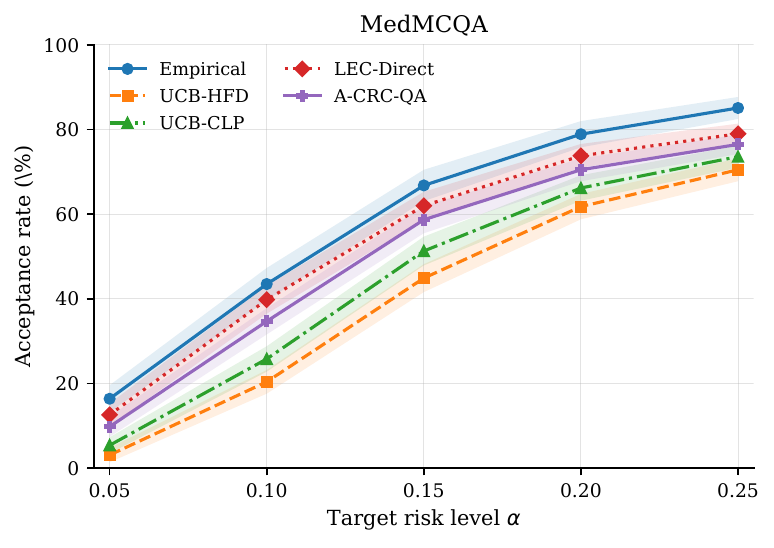}
        \caption{MedMCQA: acceptance rate.}
    \end{subfigure}

    \caption{Risk control and acceptance-rate trade-offs across
    target risk levels. Curves show the mean over repeated
    calibration--test splits, and shaded regions denote one
    standard deviation. }
    \label{fig:risk_retention}
\end{figure*}

\subsection{Performance across Target Risk Levels}
\label{sec:risk_levels}

Table~\ref{tab:risk_levels} evaluates \method{} under target
risk levels from $0.05$ to $0.25$. The reported results are
averaged over the two models using WSE on CoQA and PE on
MedMCQA.

The empirical SCER remains below the target level across all
settings. As expected, increasing $\alpha$ permits the system
to accept more answers and substantially improves power.
For example, the average acceptance rate on CoQA increases
from $11.6\%$ at $\alpha=0.05$ to $78.6\%$ at
$\alpha=0.25$.

Very small risk levels can be infeasible when the base model
does not produce a sufficiently reliable low-uncertainty
subset. At $\alpha=0.05$, the infeasibility rates are
$16\%$ on CoQA and $22\%$ on MedMCQA. This observation
is important in practice: calibration cannot create reliable
answers when the underlying model and uncertainty signal do
not support the requested operating point.

\begin{table}[t]
\centering
\caption{Performance of \method{} across target risk levels,
averaged over the two evaluated models. AR, Power, VR, and
IF are reported as percentages.}
\label{tab:risk_levels}
\resizebox{\columnwidth}{!}{
\begin{tabular}{lccrrrr}
\toprule
Dataset &  $\alpha$ & SCER
& AR  & Power  & VR  & IF  \\
\midrule
\multirow{5}{*}{CoQA}
& 0.05 & 0.043 & 11.6 & 15.9 & 12 & 16 \\
& 0.10 & 0.093 & 28.4 & 36.9 & 13 & 3 \\
& 0.15 & 0.143 & 46.2 & 56.7 & 13 & 0 \\
& 0.20 & 0.192 & 64.1 & 74.2 & 13 & 0 \\
& 0.25 & 0.241 & 78.6 & 85.5 & 14 & 0 \\
\midrule
\multirow{5}{*}{MedMCQA}
& 0.05 & 0.044 & 9.8  & 15.7 & 10 & 22 \\
& 0.10 & 0.093 & 34.7 & 52.8 & 9  & 4 \\
& 0.15 & 0.138 & 58.7 & 84.8 & 8  & 0 \\
& 0.20 & 0.188 & 70.5 & 96.0 & 9  & 0 \\
& 0.25 & 0.236 & 76.5 & 98.0 & 10 & 0 \\
\bottomrule
\end{tabular}}
\end{table}

\subsection{Robustness to Uncertainty Estimators}
\label{sec:uq_robustness}

Table~\ref{tab:uq_results} compares different uncertainty
signals at $\alpha=0.15$. All uncertainty estimators support
average risk control after calibration, showing that
\method{} does not depend on one particular definition of
uncertainty.

Nevertheless, uncertainty quality strongly affects
retention. On CoQA, WSE achieves higher AUROC and accepts
approximately four percentage points more answers than
Semantic Entropy. On MedMCQA, PE consistently outperforms
MSP. These results confirm that calibration and uncertainty
estimation play complementary roles: calibration specifies a
statistically meaningful operating point, whereas a more
discriminative uncertainty estimator increases the number of
answers that can be safely accepted.

\begin{table}[t]
\centering
\caption{Robustness of \method{} to different uncertainty
estimators at $\alpha=0.15$.}
\label{tab:uq_results}
\resizebox{\columnwidth}{!}{
\begin{tabular}{lllcccc}
\toprule
Dataset & Model & UQ
& AUROC 
& SCER 
& AR  
& Power   \\
\midrule
\multirow{4}{*}{CoQA}
& \multirow{2}{*}{LLaMA-3.1-8B}
& SemEnt & 0.691 & 0.144 & 39.1 & 49.1 \\
&
& WSE    & 0.735 & 0.145 & 42.6 & 53.4 \\
\cmidrule(lr){2-7}
& \multirow{2}{*}{Qwen2.5-7B}
& SemEnt & 0.724 & 0.142 & 46.3 & 55.6 \\
&
& WSE    & 0.762 & 0.141 & 49.8 & 59.9 \\
\midrule
\multirow{4}{*}{MedMCQA}
& \multirow{2}{*}{LLaMA-3.1-8B}
& MSP    & 0.719 & 0.140 & 51.4 & 76.4 \\
&
& PE     & 0.746 & 0.139 & 55.3 & 82.2 \\
\cmidrule(lr){2-7}
& \multirow{2}{*}{Qwen2.5-7B}
& MSP    & 0.756 & 0.138 & 58.8 & 82.6 \\
&
& PE     & 0.781 & 0.137 & 62.1 & 87.3 \\
\bottomrule
\end{tabular}}
\end{table}

\subsection{Effect of Calibration-Set Size}
\label{sec:calibration_size}

Because \method{} provides an asymptotic rather than an
exact finite-sample guarantee, calibration-set size is a
central experimental variable. We vary the number of
calibration examples from $100$ to $1{,}500$ while fixing
$\alpha=0.15$. The remaining examples are used for testing,
and results are averaged over the two models.

As shown in Table~\ref{tab:calibration_size}, small
calibration sets yield conservative thresholds and relatively
high variability. As the calibration size increases, the
finite-sample correction becomes smaller, the average SCER
approaches the target level from below, and the system
accepts more predictions. On CoQA, the acceptance rate
increases from $30.8\%$ with $100$ calibration examples to
$48.0\%$ with $1{,}500$ examples. At the same time, the
violation rate decreases from $26\%$ to $12\%$.

These trends are consistent with the asymptotic motivation
of the method. However, the nonzero violation frequencies
also confirm that the procedure should not be described as
providing conditional or high-probability finite-sample risk
control.

\begin{table}[t]
\centering
\caption{Effect of calibration-set size at $\alpha=0.15$.
SCER is shown as mean $\pm$ standard deviation over random
splits.}
\label{tab:calibration_size}
\resizebox{\columnwidth}{!}{
\begin{tabular}{lccccc}
\toprule
Dataset & $n_{\mathrm{cal}}$
& SCER
& AR 
& VR 
& IF  \\
\midrule
\multirow{5}{*}{CoQA}
& 100   & $0.127\pm0.036$ & 30.8 & 26 & 8 \\
& 250   & $0.135\pm0.027$ & 37.2 & 21 & 3 \\
& 500   & $0.139\pm0.021$ & 41.8 & 17 & 1 \\
& 1,000 & $0.143\pm0.016$ & 46.2 & 13 & 0 \\
& 1,500 & $0.145\pm0.013$ & 48.0 & 12 & 0 \\
\midrule
\multirow{5}{*}{MedMCQA}
& 100   & $0.124\pm0.032$ & 42.1 & 20 & 5 \\
& 250   & $0.131\pm0.024$ & 49.7 & 16 & 1 \\
& 500   & $0.135\pm0.019$ & 54.3 & 12 & 0 \\
& 1,000 & $0.138\pm0.014$ & 58.7 & 8  & 0 \\
& 1,500 & $0.141\pm0.012$ & 60.2 & 8  & 0 \\
\bottomrule
\end{tabular}}
\end{table}

\subsection{Ablation Study}
\label{sec:ablation}

We conduct an ablation study to isolate the effects of the
two main components of \method{}.

\begin{itemize}
    \item \textbf{Without monotonization} directly selects
    the largest threshold satisfying the corrected empirical
    constraint at that threshold.
    
    \item \textbf{Without correction} uses the monotonized
    empirical risk but removes the vanishing
    finite-sample correction.
    
    \item \textbf{Instance-wise envelope} replaces each
    non-monotone instance loss with its monotone upper
    envelope before averaging, corresponding to the direct
    finite-sample construction for general losses discussed
    in CRC~\cite{angelopoulos2024conformal}.
\end{itemize}

Table~\ref{tab:ablation} shows that removing monotonization
produces the largest acceptance rate but fails to maintain
the target risk. Removing the correction leads to average
SCER values close to or above $\alpha$ and substantially
increases the violation rate. In contrast, monotonizing each
instance-wise loss is extremely conservative, accepting fewer
than $20\%$ of predictions on both datasets. The full method
provides a more useful compromise between retention and
empirical stability.

\begin{table}[t]
\centering
\caption{Ablation study at $\alpha=0.15$, averaged over the
two evaluated models.}
\label{tab:ablation}
\resizebox{\columnwidth}{!}{
\begin{tabular}{llccc}
\toprule
Dataset & Variant
& SCER
& AR 
& VR  \\
\midrule
\multirow{4}{*}{CoQA}
& Without monotonization
& 0.164 & 55.7 & 65 \\
&
Without correction
& 0.151 & 49.4 & 41 \\
&
Instance-wise envelope
& 0.073 & 12.8 & 0 \\
&
\textbf{Full \method{}}
& \textbf{0.143}
& \textbf{46.2}
& \textbf{13} \\
\midrule
\multirow{4}{*}{MedMCQA}
& Without monotonization
& 0.171 & 69.4 & 72 \\
&
Without correction
& 0.154 & 61.5 & 46 \\
&
Instance-wise envelope
& 0.068 & 18.9 & 0 \\
&
\textbf{Full \method{}}
& \textbf{0.138}
& \textbf{58.7}
& \textbf{8} \\
\bottomrule
\end{tabular}}
\end{table}

\subsection{Discussion}
\label{sec:experimental_discussion}

The experiments provide three main observations. First,
uncalibrated uncertainty thresholds do not reliably control
the error rate among accepted answers, even when uncertainty
has reasonable error-discrimination ability. Second,
confidence-bound-based calibration provides strong
finite-sample conservativeness but may reject many correct
predictions. Third, monotonizing the empirical linear risk
reduces sensitivity to isolated feasible thresholds and
provides a practical middle ground between the aggressive
LEC-Direct rule and conservative UCB methods.

The results should nevertheless be interpreted according to
the scope of the theory. \method{} targets asymptotic
marginal risk control and does not guarantee that every
realized calibration split will satisfy the target risk.
Therefore, average SCER should always be reported together
with violation rate, infeasibility rate, and calibration-size
sensitivity. This reporting protocol prevents an average-risk
guarantee from being incorrectly presented as a
high-probability finite-sample guarantee.
\section{Conclusion}
\label{sec:conclusion}

In this paper, we studied uncertainty-aware selective question answering for large language models. Since uncertainty scores cannot perfectly distinguish correct from incorrect answers, we proposed A-CRC-QA, a post-hoc calibration framework that combines the linear expectation constraint of LEC with the asymptotic treatment of non-monotone losses in conformal risk control. The method calibrates an uncertainty threshold using held-out data and selectively accepts model answers under a user-specified target risk. 

Experiments on CoQA and MedMCQA show that the proposed framework can maintain low error rates among accepted answers while retaining more predictions than conservative confidence-bound baselines. The results also demonstrate its applicability across open-ended and closed-ended QA, different language models, and multiple uncertainty estimators. Future work will investigate finite-sample guarantees, distribution shift, and extensions to multi-model routing and human--AI collaboration.

\bibliographystyle{ACM-Reference-Format}
\bibliography{sample-base}

% \appendix









\end{document}